\documentclass{article}
\usepackage{ijcai22}

\usepackage{times}
\usepackage{soul}
\usepackage{url}
\usepackage[hidelinks]{hyperref}
\usepackage[utf8]{inputenc}
\usepackage[small]{caption}
\usepackage{graphicx}
\usepackage{amsmath}
\usepackage{amsthm}
\usepackage{booktabs}
\usepackage{algorithm}
\usepackage{algorithmic}
\title{TranSiam: Fusing Multimodal Visual Features Using Transformer for Medical Image Segmentation}

\author{
Xuejian Li$^1$
\and
Shiqiang Ma$^{1, \ast}$\and
Jijun Tang$^2$\and
Fei Guo$^{1,3}$
\affiliations
$^1$Tianjin University\\
$^2$Central South University\\
$^3$University of South Carolina, Columbia, U.S
}

\usepackage{bbding} 
\begin{document}

\maketitle

\begin{abstract}

Automatic segmentation of medical images based on multi-modality is an important topic for disease diagnosis. Although the convolutional neural network (CNN) has been proven to have excellent performance in image segmentation tasks, it is difficult to obtain global information. The lack of global information will seriously affect the accuracy of the segmentation results of the lesion area. In addition, there are visual representation differences between multimodal data of the same patient. These differences will affect the results of the automatic segmentation methods. To solve these problems, we propose a segmentation method suitable for multimodal medical images that can capture global information, named TranSiam. TranSiam is a $2$D dual path network that extracts features of different modalities. In each path, we utilize convolution to extract detailed information in low level stage, and design a ICMT block to extract global information in high level stage. ICMT block embeds convolution in the transformer, which can extract global information while retaining spatial and detailed information. Furthermore, we design a novel fusion mechanism based on cross-attention and self-attention, called TMM block, which can effectively fuse features between different modalities. On the BraTS 2019 and BraTS 2020 multimodal datasets, we have a significant improvement in accuracy over other popular methods.

\end{abstract}

$\footnote{$\ast$ equal contribution}$

\section{Introduction}

Analyzing medical images plays an important role in diagnosing diseases. However, manual segmentation of medical images relies on the experience of doctors. With the development of computer technology, it is possible to segment medical images automatically by computers. Accurate segmentation of diseased areas, human organs, infected areas from medical images can greatly improve the efficiency of diagnosing. Such as brain tumor segmentation, retinal vessels segmentation, COVID-19 segmentation and so on. Therefore, medical image segmentation has great application prospects.

\begin{figure*}[!t]
	\centerline{\includegraphics[width=7.0in]{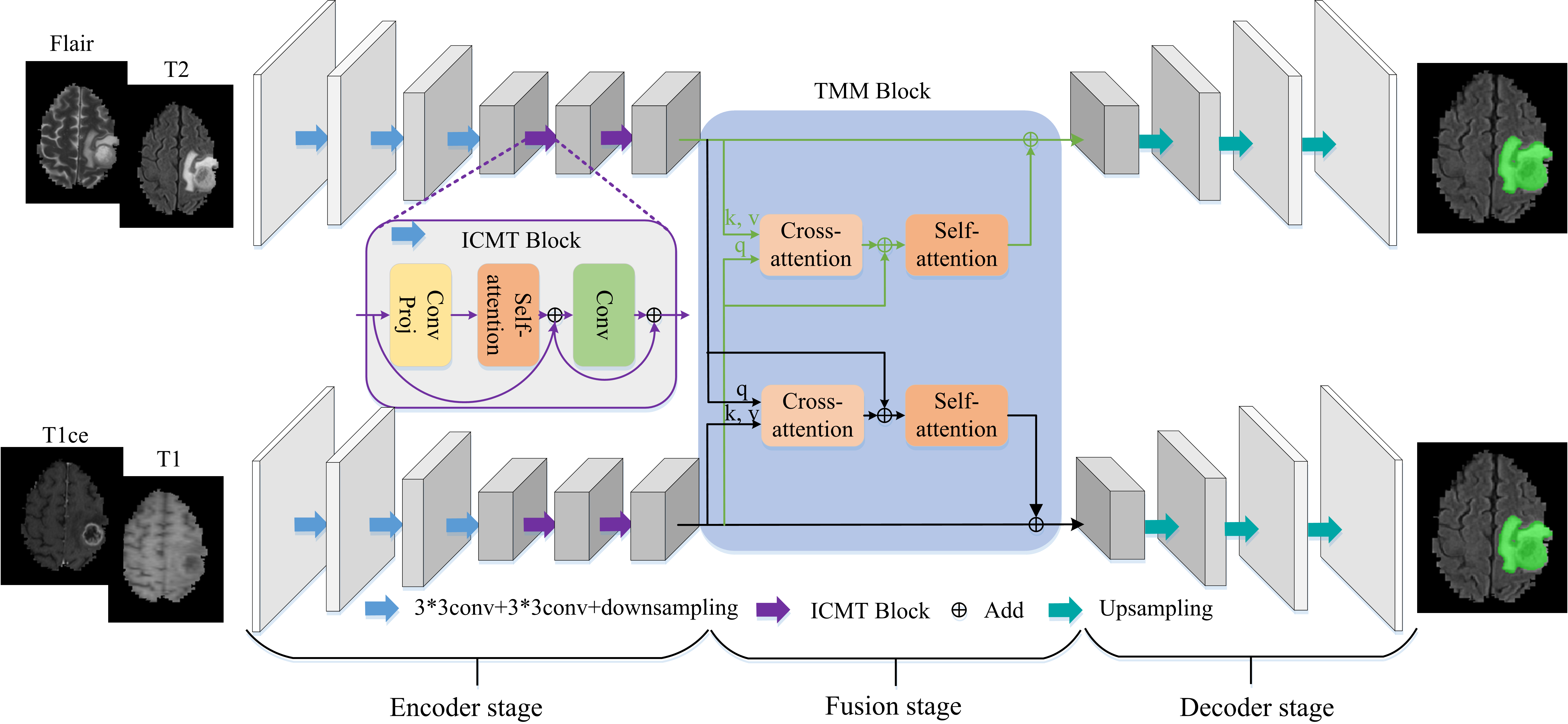}}
	\caption{The structure of TranSiam. TranSiam consists of three parts: encoder stage, fusion stage and decoder stage. ICMT block is used to capture global information, where conv proj means projection using convolution. TMM block is used to fuse multimodal features.}\label{fig:01}
\end{figure*}

In recent years, CNN has become a classic method for medical image segmentation. Such as Unet \cite{ronneberger2015u} and Nested Unet \cite{zhou2018unet++}. They had achieved satisfying results by using convolution and pooling. Because convolution has the characteristics of local receptive fields and shared weights, it can effectively extract detailed information with small amount of parameters and calculations. In addition, convolution has translation invariance, rotation invariance and so on, which makes CNN has a great generalization ability, especially for small datasets. However, the convolutional receptive field is limited, so that the global information and the correlation between distant pixels cannot be captured. Recently, with the introduction of Vit \cite{dosovitskiy2020image}, transformers \cite{vaswani2017attention} were introduced into the field of computer vision and achieved great results. Transformer showed a strong ability in capturing global information. Besides, transformer could dynamically calculate weights based on the correlation between global pixels, so it could automatically adapt to various inputs. This was beneficial for extracting high level features. However, these characteristics made transformer has a large amount of parameters, so more data was needed for training. In addition, transformer was not sensitive to detailed information. However, medical image data is less, and the image contains a lot of detailed information. Therefore, it is difficult for the transformer to be applied to medical image segmentation. To solve the above problems, we try to combine convolution and transformer to fully use their advantages. Similar to CMT \cite{guo2021cmt} and CVT \cite{wu2021cvt}, we apply depthwise convolution for replacing the linear projection of the transformer. Besides, we replace the MLP layers with standard convolution. This allows the transformer to have the characteristics of convolution, while retaining the capability of capturing global features. In addition, since low level stage contains rich detailed information and the size of feature maps are large, we use standard convolution to extract low level features. At the same time, the improved transformer is used in high level layers. In short, our method combines convolution with transformer, which greatly improves the accuracy of medical image segmentation beyond the pure transformer and CNN.

Besides, compared with single modal data, multimodal data has great feature differences. These differences will affect the results of the automatic segmentation methods. How to effectively fuse multimodal data is another difficulty for medical image segmentation. Cascaded U-Net \cite{jiang2019two} concatenated multimodal images as the input of model. pairing modal \cite{wang2020modality} exchanged the features of different modal data at each layer of model. More methods such as \cite{liu2021multimodal,prakash2021multi}. These methods cannot fully use the information between multimodal data. And the differences between them will have a negative effect. For better fusion, we design a dual path network to extract features of different modalities. To avoid the influence between multimodal data, we only merge the multimodal features in high level layers, which also reduces the complexity of the model. Besides, we propose a new fusion block, which can fully model the correlation between multimodal data and fuse multimodal features.

Our contributions can be summarized in the following points. Firstly, we make full use of the features of convolution and transformer, and combine them to improve the performance of the pure CNN and transformer for medical image segmentation. Secondly, we propose a dual path network, named TranSiam, which can effectively extract multimodal features. Thirdly, we design a new fusion block that can fully model the correlation between the multimodal features and fuse multimodal features.

\section{Related work}

Recently, transformers have shown powerful ability for extracting global information, which made up for the shortcomings of CNN. However, transformers were not easy to train with a large amount of parameters. Besides, transformers were not sensitive to detailed information. Therefore, some methods were proposed to combine CNN and transformer. For example, transunet \cite{chen2021transunet} and transbrats \cite{wang2021transbts} applied convolution and transformer for extracting low level features and global information, respectively. They effectively reduced the amount of parameters. But they used the raw transformer, which still caused the loss of some detailed information, and was affected by the positional embedding. We use convolution to change the internal structure of transformer, which can alleviate this problem. Conformer \cite{peng2021conformer} connected CNN and transformer in parallel, which could effectively merge the features of each other. Transfuse \cite{zhang2021transfuse} is similar to Conformer\cite{peng2021conformer}, but it only merged the features in the decoder stage. More related work such as \cite{guo2021cmt,wu2021cvt,dai2021coatnet} etc.

\section{Methods}

In this paper, we combine transformer and CNN to propose a new model for multimodal medical image segmentation, named TranSiam. TranSiam makes the transformer adapt well to a small number of medical image datasets. Furthermore, we combine convolution with transformer and propose ICMT block. The block can retain detailed features while capturing global features. Besides, we design a new fusion module to fuse the features of different modalities, called TMM block. It can effectively model the correlation between multimodal data. In the following, we introduce TranSiam, ICMT block and TMM block in detail, respectively.

\subsection{TranSiam}

A lot of differences exist between multimodal data. In this paper, we use the Brats $2020$ challenge dataset, which contains four modalities: T1, T1ce, Flair and T2. In our method, we treat T1, T1ce as one modality and Flair, T2 as another modality because the difference between them is small. Therefore, our TranSiam is a dual path network, which is shown as Figure~\ref{fig:01}. Compared with the single path network, the dual path network can extract the features of different modalities separately to reduce the influence between the different modalities. Our TranSiam can be divided into three parts: encoder stage, fusion stage and decoder stage.

\begin{figure}[!t]
	\centerline{\includegraphics[width=3.5in]{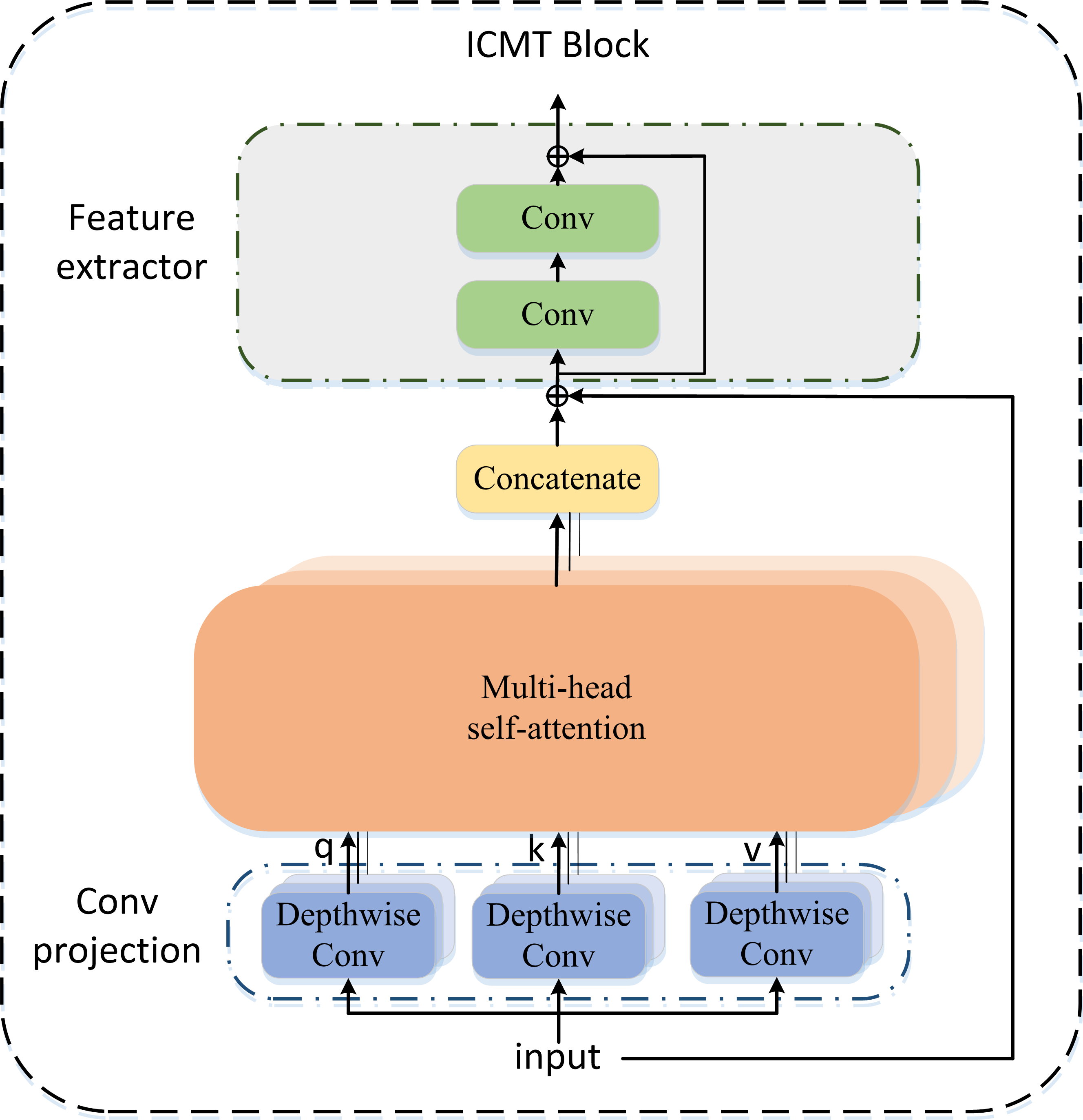}}
	\caption{The structure of ICMT Block. It includes convolutional projection, multi-head self-attention and feature extractor}\label{fig:02}
\end{figure}

Different from natural images, medical images contain a large amount of detailed information, which can seriously affect the accuracy of image segmentation. So we apply convolution for extracting low level features in the encoder stage. The mechanism of convolutional local windows can effectively extract these detailed information. In addition, the characteristics of convolutional shared weight can reduce the number of parameters of our model. Then we use convolution with stride $2$ to down sample, which can filter out redundant information and reduce the loss of detailed information. After that, we design ICMT block to extract high level features. ICMT block integrates the characteristics of convolution and transformer, which can model the correlation between distant pixels while maintaining detailed information. It is beneficial to obtain the shape of the lesion area and the information of its relative position. In the fusion stage, we propose a novel fusion block, named TMM block. The block can effectively utilize the information between different modalities by establishing the correlation between them. In the decoder stage, we apply deconvolution for restoring the fused features to the same size as the input so that our TranSiam can achieve end-to-end segmentation. Finally, the average of the two outputs is used as the final result.

\subsection{ICMT block}

\begin{figure}[!t]
	\centerline{\includegraphics[width=3.5in]{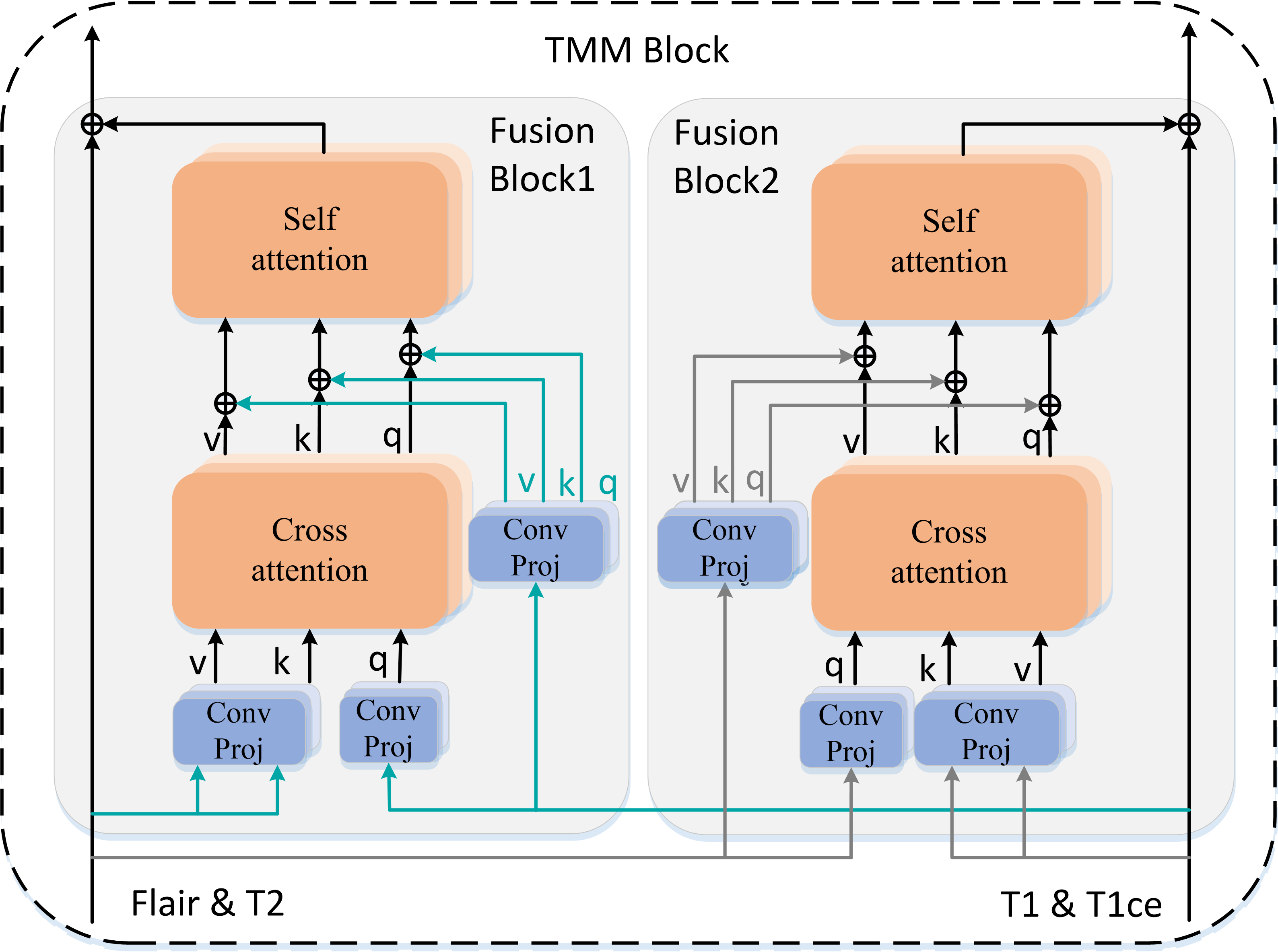}}
	\caption{The structure of TMM Block. It consists of fusion block1 and fusion block2. All fusion blocks include convolutional projection, cross-attention and self-attention.}\label{fig:03}
\end{figure}

We design ICMT Block to obtain global information while preserving detailed and spatial information. The structure of ICMT Block is shown as Figure~\ref{fig:02}. Similar to \cite{guo2021cmt,wu2021cvt}, we use depthwise convolution to embed features instead of the linear projection of the original transformer. Compared with linear projection, convolutional projection can preserve the spatial position information of pixels, and convolutional projection does not require position embedding. It is very beneficial for medical image segmentation, because the location information of the lesion relative to the organ is very important. Besides, convolutional projection can reduce the loss of information. After that, multi-head self-attention is used to establish correlation between distant pixels, which can obtain global information. Finally, we use standard convolution to replace the MLP layer of the original transformer, which can fully extract spatial information. The ICMT block can be formulated as follows:
\begin{equation}
	q=xw_{q}, k=xw_{k}, v=xw_{v}
\end{equation}%
\begin{equation}
	Att=Softmax(\dfrac{qk^{T}}{\sqrt{d_{k}}})v+x
\end{equation}%
\begin{equation}
	y=Conv(Att)+Att
\end{equation}%

where, $x$ and $y$ represent the input and output of the ICMT block, respectively, $w_{i}$ ($i=[q,k,v]$) represents the weight of the corresponding depthwise convolution, $q$, $k$, $v$ represent query, key, value of the transformer, respectively, Conv represents a $3 \times 3$ convolution with stride of $1$, and Att represents the output of self-attention. 

\subsection{TMM block}

\begin{table}[!t]
 \caption{Comparing TMM block with other fusion methods on BraTS 2020 validation set. We show the Dice (\%), Sensitivity (\%), Specificity (\%) and HD95 of each model in the whole brain tumor segmentation. Without means no fusion method is used.}
 \centering
 \resizebox{1\columnwidth}{!}{
 \begin{tabular}{ccccc}
  \hline
  Fusion  & Dice & Sensitivity & Specificity & HD95  \\
  \hline
  Without     & 88.53  & 88.60 & 99.89 & 6.436 \\
  Add         & 88.86  & 88.09 & 99.90 & \textbf{5.041} \\
  Concatenate    & 89.05  & 88.44 & 99.91 & 5.094 \\
  Attention gate   & 88.81  & 88.51 & 99.90 & 5.562 \\
  TMM block        & \textbf{89.34}  & \textbf{88.88} & \textbf{99.91} & 5.655 \\
  \hline
 \end{tabular}
 }
 \label{tab:TMM}
\end{table}

\begin{table}[!t]
 \caption{Comparison TranSiam with and without ICMT Block on BraTS 2020 validation set. We show the Dice (\%), Sensitivity (\%), Specificity (\%) and HD95 of each model in the whole brain tumor segmentation. w/o represents without, w/ means with.}
 \centering
  \resizebox{1\columnwidth}{!}{
 \begin{tabular}{ccccc}
  \hline
  Model  & Dice & Sensitivity & Specificity & HD95  \\
  \hline
  TranSiam w/o ICMT  & 88.60  & 87.61 & 99.91 & 5.880 \\
  TranSiam w/ CMT  & 88.99  & 88.29 & 99.90 & \textbf{5.408} \\
  TranSiam w/ ICMT   & \textbf{89.34}  & \textbf{88.88} & \textbf{99.91} & 5.655 \\
  \hline
 \end{tabular}
 }
 \label{tab:ICMT}
\end{table}

\begin{table}[!t]
 \caption{Ablation experiment of key components inside ICMT block on BraTS 2020 validation set. We show the Dice (\%), Sensitivity (\%) of each model in the whole brain tumor segmentation. Conv means convolution.}
 \centering
 \resizebox{1\columnwidth}{!}{
 \begin{tabular}{cccc}
  \hline
  Conv projection  & Conv instead of MLP & Dice & Sensitivity  \\
  \hline
   &  & 88.74 & 87.91\\
  $\surd$ &   & 89.06 & 88.50 \\
   & $\surd$ & 89.06 & 88.22 \\
  $\surd$ & $\surd$ & \textbf{89.34} & \textbf{88.88} \\
  \hline
 \end{tabular}
 }
 \label{tab:ICMT ablation}
\end{table}

\begin{table}[!t]
 \caption{Comparison of single path and dual path on BraTS 2020 validation set. We show the Dice (\%), Sensitivity (\%), Specificity (\%) and HD95 of each model in the whole brain tumor segmentation.}
 \centering
  \resizebox{1\columnwidth}{!}{
 \begin{tabular}{ccccc}
  \hline
  Model  & Dice & Sensitivity & Specificity & HD95  \\
  \hline
  Single path  & 88.65  & 87.22 & 99.91 & \textbf{5.255} \\
  Dual path   & \textbf{89.34}  & \textbf{88.88} & \textbf{99.91} & 5.655 \\
  \hline
 \end{tabular}
 }
 \label{tab:single and dual path}
\end{table}

Using complementary information between multimodal data can greatly improve the accuracy of segmentation. Therefore, we propose a new fusion mechanism, called TMM block, which is used to fuse high level features of different modalities. As shown as Figure~\ref{fig:03}, the TMM block is composed of two fusion blocks. Fusion block1 fuses the features of T1 and T1ce into Flair and T2. Fusion block2 fuses Flair and T2 into T1 and T1ce. These fusion blocks consist of cross-attention \cite{vaswani2017attention} and self-attention. We take fusion block1 as an example to describe.

Firstly, we map the features of T1 and T1ce into q (query) vectors and the features of Flair and T2 into k (key) and v (value) vectors using depthwise convolution. Then we use cross-attention to calculate the similarity between q and k, and then use the similarity to weight v. In other words, cross-attention highlights similar features between two modalities by establishing the correlation between them. Since these features can be extracted from both modal data, they should be set with high confidence. In addition, T1 and T1ce contain some unique features, so we add the features of T1 and T1ce to Flair and T2 by add. Next, we apply self-attention for establishing the correlation between global pixels and fully fusing the features between the two modalities. Finally, the fused features are added to Flair and T2 to assist the segmentation. The fusion block1 can be formulated as follows:

\begin{equation}
	q_{1}=x_{2}w_{q1}, k_{1}=x_{1}w_{k1}, v_{1}=x_{1}w_{v1}
\end{equation}%
\begin{equation}
	Cross_{att}=Softmax(\dfrac{q_{1}k_{1}^{T}}{\sqrt{d_{k_{1}}}})v_{1}
\end{equation}%
\begin{equation}
	f=Cross_{att}+x_{2}
\end{equation}%
\begin{equation}
	q_{2}=fw_{q2}, k_{2}=fw_{k2}, v_{2}=fw_{v2}
\end{equation}%
\begin{equation}
	y=Conv(Softmax(\dfrac{q_{2}k_{2}^{T}}{\sqrt{d_{k_{2}}}})v_{2})+x_{1}
\end{equation}%
where, $x_{1}$ represents the features of Flair and T2, $x_{2}$ represents the features of T1 and T1ce, y represents the output of fusion block1, and $Cross_{att}$ means cross-attention.

\subsection{Loss function}
We use a joint loss to improve the training efficiency of our network and the accuracy of segmentation. The joint loss consists of multi-class cross-entropy loss and Dice loss. Dice loss was first proposed in \cite{1}, which focused on the prediction results of foreground area and reduced the effect of class imbalance. But Dice loss will be difficult to converge, when the foreground area is too small. Therefore, we combine the multi-class cross-entropy loss $L_{CE}$ with the Dice loss $L_{dice}$ to alleviate the difficulty. The joint loss can be formulated as follows:

\begin{equation}
	L_{joint}=\alpha_{1}L_{CE}+\alpha_{2}L_{dice}
\end{equation}%
\begin{equation}
	L_{dice}=\frac{2|y \cap \hat{y}|}{|y|+|\hat{y}|}
\end{equation}%
\begin{equation}
	L_{CE}=-\sum_{n=1}^N\sum_{k=1}^K(y_{nk}ln\hat{y}_{nk}+(1-y_{nk})ln(1-\hat{y}_{nk}))
\end{equation}%
where, the target is an $N \times K$ matrix. $y$ is the ground truth. $\hat{y}$ refers to the prediction. $alpha_{1}$ and $alpha_{2}$ are the loss weights of the multi-class cross-entropy loss and Dice loss. $alpha_{1}$ and $alpha_{2}$ are set to $0.3$ and $0.7$, respectively.

\section{Experiments and results}

\subsection{Dataset}

We evaluate the performance of TranSiam on the BraTS $2019$ and BraTS $2020$ Challenge dataset. The BraTS $2020$ training dataset contains $369$ $3D$ brain MR images. The BraTS $2019$ training dataset consists of $335$ $3$D brain MR images. Both BraTS $2019$ and $2020$ validation sets contain $125$ $3$D brain MR images. We upload the segmentation results to the online evaluation system for obtaining evaluation results on the validation dataset. The dimensions of the $3$D brain images of all patients were $240 \times 240 \times 155$. Both training and validation datasets contain glioblastoma (GBM/HGG) and lower grade glioma (LGG). Lesion areas in brain tumor patients are classified as edematous, enhancing tumor, and necrotic. These ground truth are annotated by expert. Each brain MR image includes four modalities: T1, T1ce, Flair and T2. We cropped all brain data to remove useless background regions. The size of the cropped image is $176\times144\times155$.

\begin{table}[!t]
\caption{Comparison with the classic methods on BraTS 2020 validation set. We show the Dice (\%), Sensitivity(\%), HD95 and Parameters(M) of each model in the whole brain tumor segmentation.}
\centering
\resizebox{1\columnwidth}{!}{
\begin{tabular}{ccccc}
\hline
Model  & Dice & Sensitivity & HD95 & Params \\
\hline
\hline
U-Net       & 88.63  & 86.91  & 5.49  & {\bf 7.77} \\
Nested U-Net       & 88.86  & 87.43  & 6.44 & 9.16 \\
Attention U-Net       & 88.67  & 86.82  & 5.46  & 7.86\\
CENet   & 88.98  & 87.23  & {\bf 5.14}  & 9.18\\
TransBTS w/o TTA & 89.00 & - & 6.47 & 32.99 \\
TranSiam   & {\bf 89.34}  & {\bf 88.88}  & 5.65  & 7.98\\
\hline
\end{tabular}
}
\label{tab:5}
\end{table}

\begin{table}[!t]
\caption{Comparison with the classic methods on BraTS 2019 validation set. We show the Dice (\%), Sensitivity (\%), Specificity (\%) and HD95 of each model in the whole brain tumor segmentation.}
\centering
\resizebox{1\columnwidth}{!}{
\begin{tabular}{ccccc}
\hline
Model  & Dice & HD95 & Sensitivity & Specificity \\
\hline
\hline
3D U-Net       & 87.38  &  9.43  & - & - \\
V-Net       & 88.73  & 6.26  & - & - \\
KiU-Net       & 87.60  & 8.94  & - & - \\
Attention U-Net   & 88.81  & 7.76   & - & - \\
Li et al.     & 88.60  & 6.23   & - & - \\
TransBTS w/o TTA   & 88.89  & 7.60   & - & - \\
TranSiam   & {\bf 89.26}  & {\bf 4.77}   & 88.52  & 99.42  \\
\hline
\end{tabular}
}
\label{tab:7}
\end{table}

\subsection{Evaluation metrics}
We employ four conventional metrics to quantitatively evaluate the segmentation performance of TranSiam. They are Dice similarity coefficient (\%), Specificity(\%), Sensitivity(\%) and Hausdorff distance (HD95) respectively. Dice similarity coefficient calculates the volume overlap between the prediction mask and the ground truth. Sensitivity and specificity are statistical measures of the performance of binary classification tests. Hausdorff distance calculates the distance between the prediction mask and the ground truth in metric space. Hausdorff distance is more sensitive to outliers.

\subsection{Experiment details}
Our experiments are mainly trained on an NVIDIA V100. All models are trained for $100$ epochs. Batch size is set to $72$. The SGD is used as our optimizer. The initial learning rate is set to $0.03$, the weight decay is $0.0001$ and the momentum is $0.9$. We use conventional strategies for image augmentation including random flip, random rotate, random scale and random crop. The probability of each strategy being used is $0.5$. In the test phase, we use the weights of the last $5$ epochs to make predictions on the test set separately. We take the average of $5$ results as the final result to enhance the generalization ability of our model.

\begin{figure*}[!t]
	\centerline{\includegraphics[width=7.0in]{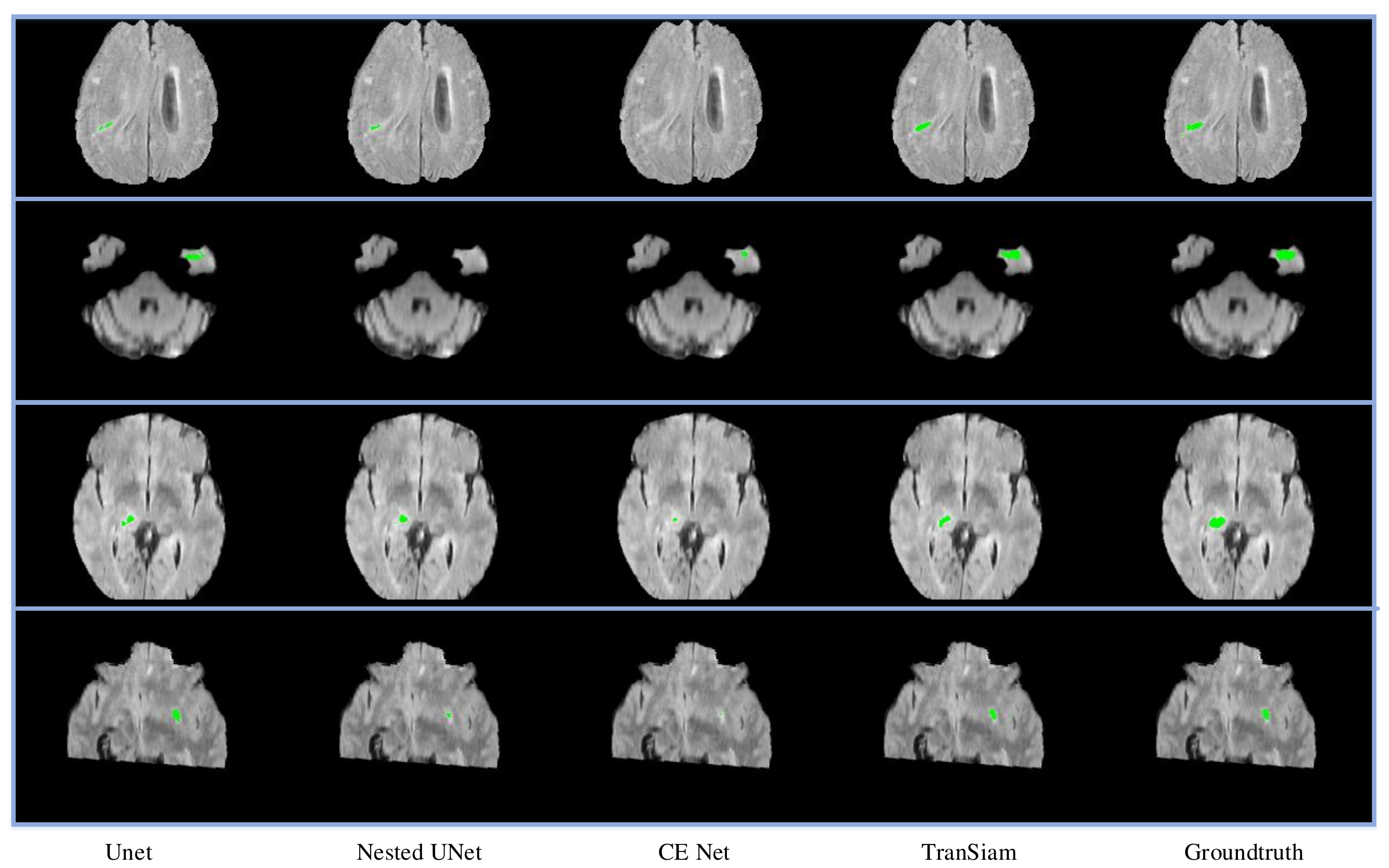}}
	\caption{Comparison of TranSiam with other classical methods. From left to right: Unet, Nested UNet, CE Net, TranSiam, Groundtruth. Whole tumors are marked in green.}\label{fig:compare}
\end{figure*}

\subsection{Ablation study}
In this section, we evaluate the contribution of the TMM block for multimodal feature fusion. There are great differences in the visual features of different modalities, but there are few effective fusion methods to fuse multimodal features. We compare several feature fusion methods with the TMM block to objectively evaluate the performance of our TMM block. The experimental results are shown in Table \ref{tab:TMM}. The TMM block achieves best results than other fusion methods in Dice, Specificity, and Sensitivity, which indicates our TMM block is meaningful and can segment more lesions.

Then, we evaluate the effect of ICMT block on the segmentation performance of TranSiam, which is shown as Table \ref{tab:ICMT}. We replace TranSiam's ICMT block with the convolution block. In addition, we also compare ICMT block with CMT block. TranSiam with ICMT block is significantly improved compared to without ICMT block and with CMT block. It proves that ICMT block combines the advantages of convolution and transformer is important. It establishes the correlation between distant pixels while maintaining the detailed information.

We conduct ablation experiments to evaluate the effectiveness of each key component in our proposed ICMT block. The experimental results are shown in Table \ref{tab:ICMT ablation}. Key components of the ICMT block include Conv projection and Feature extractor. We replace the linear projection and MLP layer of the raw transformer with depthwise convolution and residual block with $3\times3$ convolution. The experimental results show that the results are the worst using linear projection and MLP layer. Replacing the linear projection with the Conv projection and replacing the MLP layer with a $3\times3$ residual convolution block achieves similar results. Using both Conv projection and $3\times3$ residual convolution block get the best results. The experimental results prove that we embed the convolution inside the transformer is meaningful.

Finally, we evaluate the effectiveness of the Key Components of the proposed framework. We test the effect of the dual path architecture on the segmentation performance of TranSiam. As shown in Table \ref{tab:single and dual path}. The baseline network uses single path architecture. We concatenate the images of the four modalities as its input. Compared to the baseline network, TranSiam achieves the best results in all four metrics. Experimental results show that TranSiam can more fully utilize the complex visual features of multimodal data.

\subsection{Comparison with classic Methods}
Firstly, we compare TranSiam with classic methods for medical image segmentation on BraTS $2020$ validation set, which include U-Net, Nested UNet, Attention U-Net and CENet. We reproduce the above methods under the same benchmark. TranSiam does not use ensemble learning and special post-processing for a fair comparison. Besides, we upload the segmentation results of all methods to the online evaluation system. So our results are credible.The experimental results are shown in Table \ref{tab:5}. Our method has satisfying segmentation performance and achieves best results in Dice and Sensitivity. It is worth noting that our framework outperforms other methods in Sensitivity, i.e., our method is able to segment more lesion areas. Besides, TransBTS also combines transformer and convolution. But TransBTS did not pay attention to the difference between multimodal data and the problem that the transformer loses detailed information. Therefore, the evaluation results of our proposed $2$D TranSiam outperform the $3$D TransBTS. In addition, the number of parameters of TranSiam (only $7.98$M) is much smaller than that of TransBTS ($32.99$M).

Finally, we compare TranSiam with $3$D classic methods for medical image segmentation on BraTS $2019$ validation set, which include $3$D U-Net \cite{2}, V-Net \cite{1}, KiU-Net \cite{3}, Attention U-Net \cite{4}, Li et al. \cite{5} and TransBTS \cite{wang2021transbts}. The experimental results are shown in Table \ref{tab:7}. Our TranSiam achieves an average Dice similarity score of $89.26\%$ on the whole brain tumor segmentation task, which is higher than the results of the previously proposed classic $3$D methods. The average HD95 of our TranSiam with $4.77$ is also higher than other methods. Notably, our method does not use post-processing. Therefore, the experimental results of our method are all obtained by uploading the output of TranSiam directly to the online evaluation system.

\subsection{Visual analysis}

We perform a visual analysis for our experimental results. It is worth noting that the validation set of brats 2019 and brats 2020 does not provide groundtruth. Therefore, we apply the brats 2020 training set for training our model, and select part of the brats 2019 training set as our validation set. On the validation set, we visualize the segmentation results of several classical networks, which is shown as Figure~\ref{fig:compare}. We use green to mark whole tumor. As can be seen from Figure~\ref{fig:compare}, although our TranSiam integrates the transformer, it can still segment small tumor. It means that we combine convolution with transformer is meaningful. It can alleviate the problem that the transformer loses detailed information. With the help of the transformer, TranSiam is even able to segment tumors that are not segmented by traditional CNNs. It means that our TranSiam fully combines the advantages of transformers and convolutions for multimodal medical image segmentation. 

\section{Conclusion}

In this paper, a dual path network is proposed for multimodal medical image segmentation. It combines the two mainstream models of CNN and transformer to improve the ability of extracting local information and global information. Besides, we combine cross-attention and self-attention and propose a TMM block to fuse multimodal features, which can effectively find similar features between different modalities and use the unique features of each modality. Importantly, extensive ablation experiments and comparative experiments prove that TranSiam can achieve better results than classical CNN and original transformer.

\bibliographystyle{named}
\bibliography{ijcai22}

\begin{thebibliography}{}

\bibitem[\protect\citeauthoryear{Chen \bgroup \em et al.\egroup
  }{2021}]{chen2021transunet}
Jieneng Chen, Yongyi Lu, Qihang Yu, Xiangde Luo, Ehsan Adeli, Yan Wang, Le~Lu,
  Alan~L Yuille, and Yuyin Zhou.
\newblock Transunet: Transformers make strong encoders for medical image
  segmentation.
\newblock {\em arXiv preprint arXiv:2102.04306}, 2021.

\bibitem[\protect\citeauthoryear{{\c{C}}i{\c{c}}ek \bgroup \em et al.\egroup
  }{2016}]{2}
{\"O}zg{\"u}n {\c{C}}i{\c{c}}ek, Ahmed Abdulkadir, Soeren~S. Lienkamp, Thomas
  Brox, and Olaf Ronneberger.
\newblock 3d u-net: Learning dense volumetric segmentation from sparse
  annotation.
\newblock In Sebastien Ourselin, Leo Joskowicz, Mert~R. Sabuncu, Gozde Unal,
  and William Wells, editors, {\em Medical Image Computing and
  Computer-Assisted Intervention -- MICCAI 2016}, pages 424--432, Cham, 2016.
  Springer International Publishing.

\bibitem[\protect\citeauthoryear{Dai \bgroup \em et al.\egroup
  }{2021}]{dai2021coatnet}
Zihang Dai, Hanxiao Liu, Quoc~V Le, and Mingxing Tan.
\newblock Coatnet: Marrying convolution and attention for all data sizes.
\newblock {\em arXiv preprint arXiv:2106.04803}, 2021.

\bibitem[\protect\citeauthoryear{Dosovitskiy \bgroup \em et al.\egroup
  }{2020}]{dosovitskiy2020image}
Alexey Dosovitskiy, Lucas Beyer, Alexander Kolesnikov, Dirk Weissenborn,
  Xiaohua Zhai, Thomas Unterthiner, Mostafa Dehghani, Matthias Minderer, Georg
  Heigold, Sylvain Gelly, et~al.
\newblock An image is worth 16x16 words: Transformers for image recognition at
  scale.
\newblock {\em arXiv preprint arXiv:2010.11929}, 2020.

\bibitem[\protect\citeauthoryear{Guo \bgroup \em et al.\egroup
  }{2021}]{guo2021cmt}
Jianyuan Guo, Kai Han, Han Wu, Chang Xu, Yehui Tang, Chunjing Xu, and Yunhe
  Wang.
\newblock Cmt: Convolutional neural networks meet vision transformers.
\newblock {\em arXiv preprint arXiv:2107.06263}, 2021.

\bibitem[\protect\citeauthoryear{Jiang \bgroup \em et al.\egroup
  }{2019}]{jiang2019two}
Zeyu Jiang, Changxing Ding, Minfeng Liu, and Dacheng Tao.
\newblock Two-stage cascaded u-net: 1st place solution to brats challenge 2019
  segmentation task.
\newblock In {\em International MICCAI Brainlesion Workshop}, pages 231--241.
  Springer, 2019.

\bibitem[\protect\citeauthoryear{Li \bgroup \em et al.\egroup }{2019}]{5}
Xiangyu Li, Gongning Luo, and Kuanquan Wang.
\newblock Multi-step cascaded networks for brain tumor segmentation.
\newblock In {\em International MICCAI Brainlesion Workshop}, pages 163--173.
  Springer, 2019.

\bibitem[\protect\citeauthoryear{Liu \bgroup \em et al.\egroup
  }{2021}]{liu2021multimodal}
Yicheng Liu, Jinghuai Zhang, Liangji Fang, Qinhong Jiang, and Bolei Zhou.
\newblock Multimodal motion prediction with stacked transformers.
\newblock In {\em Proceedings of the IEEE/CVF Conference on Computer Vision and
  Pattern Recognition}, pages 7577--7586, 2021.

\bibitem[\protect\citeauthoryear{Milletari \bgroup \em et al.\egroup
  }{2016}]{1}
Fausto Milletari, Nassir Navab, and Seyed-Ahmad Ahmadi.
\newblock V-net: Fully convolutional neural networks for volumetric medical
  image segmentation.
\newblock In {\em 2016 Fourth International Conference on 3D Vision (3DV)},
  pages 565--571, 2016.

\bibitem[\protect\citeauthoryear{Oktay \bgroup \em et al.\egroup }{2018}]{4}
Ozan Oktay, Jo~Schlemper, Loic~Le Folgoc, Matthew Lee, Mattias Heinrich,
  Kazunari Misawa, Kensaku Mori, Steven McDonagh, Nils~Y Hammerla, Bernhard
  Kainz, et~al.
\newblock Attention u-net: Learning where to look for the pancreas.
\newblock {\em arXiv preprint arXiv:1804.03999}, 2018.

\bibitem[\protect\citeauthoryear{Peng \bgroup \em et al.\egroup
  }{2021}]{peng2021conformer}
Zhiliang Peng, Wei Huang, Shanzhi Gu, Lingxi Xie, Yaowei Wang, Jianbin Jiao,
  and Qixiang Ye.
\newblock Conformer: Local features coupling global representations for visual
  recognition.
\newblock {\em arXiv preprint arXiv:2105.03889}, 2021.

\bibitem[\protect\citeauthoryear{Prakash \bgroup \em et al.\egroup
  }{2021}]{prakash2021multi}
Aditya Prakash, Kashyap Chitta, and Andreas Geiger.
\newblock Multi-modal fusion transformer for end-to-end autonomous driving.
\newblock In {\em Proceedings of the IEEE/CVF Conference on Computer Vision and
  Pattern Recognition}, pages 7077--7087, 2021.

\bibitem[\protect\citeauthoryear{Ronneberger \bgroup \em et al.\egroup
  }{2015}]{ronneberger2015u}
Olaf Ronneberger, Philipp Fischer, and Thomas Brox.
\newblock U-net: Convolutional networks for biomedical image segmentation.
\newblock In {\em International Conference on Medical image computing and
  computer-assisted intervention}, pages 234--241. Springer, 2015.

\bibitem[\protect\citeauthoryear{Valanarasu \bgroup \em et al.\egroup
  }{2021}]{3}
Jeya Maria~Jose Valanarasu, Vishwanath~A. Sindagi, Ilker Hacihaliloglu, and
  Vishal~M. Patel.
\newblock Kiu-net: Overcomplete convolutional architectures for biomedical
  image and volumetric segmentation.
\newblock {\em IEEE Transactions on Medical Imaging}, pages 1--1, 2021.

\bibitem[\protect\citeauthoryear{Vaswani \bgroup \em et al.\egroup
  }{2017}]{vaswani2017attention}
Ashish Vaswani, Noam Shazeer, Niki Parmar, Jakob Uszkoreit, Llion Jones,
  Aidan~N Gomez, {\L}ukasz Kaiser, and Illia Polosukhin.
\newblock Attention is all you need.
\newblock In {\em Advances in neural information processing systems}, pages
  5998--6008, 2017.

\bibitem[\protect\citeauthoryear{Wang \bgroup \em et al.\egroup
  }{2020}]{wang2020modality}
Yixin Wang, Yao Zhang, Feng Hou, Yang Liu, Jiang Tian, Cheng Zhong, Yang Zhang,
  and Zhiqiang He.
\newblock Modality-pairing learning for brain tumor segmentation.
\newblock {\em arXiv preprint arXiv:2010.09277}, 2020.

\bibitem[\protect\citeauthoryear{Wang \bgroup \em et al.\egroup
  }{2021}]{wang2021transbts}
Wenxuan Wang, Chen Chen, Meng Ding, Hong Yu, Sen Zha, and Jiangyun Li.
\newblock Transbts: Multimodal brain tumor segmentation using transformer.
\newblock In {\em International Conference on Medical Image Computing and
  Computer-Assisted Intervention}, pages 109--119. Springer, 2021.

\bibitem[\protect\citeauthoryear{Wu \bgroup \em et al.\egroup
  }{2021}]{wu2021cvt}
Haiping Wu, Bin Xiao, Noel Codella, Mengchen Liu, Xiyang Dai, Lu~Yuan, and Lei
  Zhang.
\newblock Cvt: Introducing convolutions to vision transformers.
\newblock {\em arXiv preprint arXiv:2103.15808}, 2021.

\bibitem[\protect\citeauthoryear{Zhang \bgroup \em et al.\egroup
  }{2021}]{zhang2021transfuse}
Yundong Zhang, Huiye Liu, and Qiang Hu.
\newblock Transfuse: Fusing transformers and cnns for medical image
  segmentation.
\newblock {\em arXiv preprint arXiv:2102.08005}, 2021.

\bibitem[\protect\citeauthoryear{Zhou \bgroup \em et al.\egroup
  }{2018}]{zhou2018unet++}
Zongwei Zhou, Md~Mahfuzur~Rahman Siddiquee, Nima Tajbakhsh, and Jianming Liang.
\newblock Unet++: A nested u-net architecture for medical image segmentation.
\newblock In {\em Deep learning in medical image analysis and multimodal
  learning for clinical decision support}, pages 3--11. Springer, 2018.

\end{thebibliography}

\end{document}